\documentclass[runningheads]{llncs}
\usepackage[T1]{fontenc}
\usepackage{xcolor}
\usepackage{multirow}
\usepackage{amsmath}
\usepackage{amsfonts}
\usepackage{diagbox}
\PassOptionsToPackage{hyphens}{url}\usepackage{hyperref}
\usepackage{graphicx}
\begin{document}
%
\title{The Adversarial AI-Art: Understanding, Generation, Detection, and Benchmarking}
%
\author{Yuying Li\inst{1} \and
Zeyan Liu\inst{1} \and
Junyi Zhao\inst{1} \and
Liangqin Ren\inst{1} \and \\
Fengjun Li\inst{1} \and
Jiebo Luo\inst{2}\and
Bo Luo\inst{1}}
%
%
\institute{EECS/I2S The University of Kansas, Lawrence, KS, USA \and Department of Computer Science, University of Rochester,
Rochester, NY 14627 \\
\email{\{yuyingli, zyliu, junyi.zhao, liangqinren, fli, bluo\}@ku.edu}, \email{jluo@cs.rochester.edu}
}
\maketitle              
\begin{abstract}
Generative AI models can produce high-quality images based on text prompts. The generated images often appear indistinguishable from images generated by conventional optical photography devices or created by human artists (i.e., real images). While the outstanding performance of such generative models is generally well received, security concerns arise. For instance, such image generators could be used to facilitate fraud or scam schemes, generate and spread misinformation, or produce fabricated artworks. In this paper, we present a systematic attempt at understanding and detecting AI-generated images (AI-art) in adversarial scenarios. First, we collect and share a dataset of real images and their corresponding artificial counterparts generated by four popular AI image generators. The dataset, named ARIA, contains over 140K images in five categories: artworks (painting), social media images, news photos, disaster scenes, and anime pictures. This dataset can be used as a foundation to support future research on adversarial AI-art. Next, we present a user study that employs the ARIA dataset to evaluate if real-world users can distinguish with or without reference images.  In a benchmarking study, we further evaluate if state-of-the-art open-source and commercial AI image detectors can effectively identify the images in the ARIA dataset.  Finally, we present a ResNet-50 classifier and evaluate its accuracy and transferability on the ARIA dataset.

\end{abstract}
\section{Introduction}
The rise of artificial intelligence has been rapidly reshaping the field of multimedia content creation in the past two years. AI companies like OpenAI, Midjourney, and StarryAI have developed tools that are highly accessible and user-friendly for the general public. As of September 2023, Midjourney is reported to have over 16 million users \cite{10million}. Without any expertise in AI or art, they can create high-quality images simply by supplying simple descriptive words as prompts. What once would require hours and or days for the photographers and artists can now be imitated and replicated within moments. 




As AI technologies continue to blur the boundaries between human and machine creativity, broad concerns and controversies over creativity, ethics, and integrity have emerged \cite{epstein2020gets,wong2024ai}. 
First, the widespread adoption of AI-art has significant impacts on copyright and authorship. Notably, online communities such as Newgrounds and FurAffinity have banned AI-generated content from their platforms \cite{banai}. Art competitions also enforce restrictions after rising controversies over AI-art \cite{roose2022ai}. According to \cite{survey_ethics}, 74\% of the artists believe AI-art is unethical, and there is reportedly a growing public interest among artists to combat unauthorized image usage by AI companies \cite{lawsuit}. Meanwhile, this issue also influences legal frameworks. For example, the United States Copyright Office has ruled that works incorporating AI-generated content must demonstrate human authorship to qualify for copyright registration \cite{usco}. 

Moreover, the proliferation of AI-based image generation poses substantial security risks, as these realistically rendered images can be weaponized to spread disinformation or commit fraud. For example, the ease of creating convincing articles with AI-generated visuals has led to a more than tenfold increase in fake news websites \cite{aifakenews1000}. Many threat actors have been reported to use AI to mislead public perception related to elections \cite{aielection1,aielection2}, which results in urgent calls for enhanced governance and mitigation \cite{aigovern2}. Another example of security risk is scamming. A rapid growth in AI-powered fake profiles has been witnessed on social media \cite{deepfakesocial} and dating apps \cite{deepfakedating}. These scenarios underscore the need to verify the authenticity of digital media and raise public awareness of AI misuse.


In this paper, we are motivated by the following questions: What are the practical scenarios in which adversarial actors exploit AI-art for malicious purposes? Can human examiners identify them with or without references? Are there any automatic tools to reliably detect AI-generated images and mitigate these risks? We investigate three primary scenarios where adversarial AI art makes a significant impact: (1) social media fraud, (2) fake news and misinformation, and (3) unauthorized art style imitation. We create images in five distinct categories using four leading AI image generation platforms, utilizing two modes of generation: generation from only text prompts and generation from text prompts and human image seeds. We collect and share the \underline{A}dversa\underline{RI}al \underline{A}I-Art (ARIA) dataset with over 140K images, including 127K adversarial AI images. 

With the ARIA dataset, we carry out a user study to assess human judgment in identifying AI-art. With 4,720 annotations from 472 participants, the average accuracy rates were 68.00\% and 65.24\% for users with and without references, respectively, indicating the ineffectiveness of manual inspection. We analyze the factors that may affect human identification, such as the image content, the generator, and the users' domain expertise. We further benchmark nine state-of-the-art AIGC detectors in the research literature and five online/commercial detection services. Almost all of the detectors provide unsatisfactory performance, especially on samples generated with mixed prompts of images and text. Finally, we discover that supervised classifiers trained on our ARIA dataset are more effective. In particular, models trained using images from Midjourney show better generalization capabilities across different generation platforms.


In summary, our main contributions are summarized as follows: 
\vspace{-2mm}

\begin{enumerate}
\item We make a systematic attempt at understanding and detecting adversarial AI-Art. 
\item We collect and share the first comprehensive adversarial AI-art dataset with paired human-generated images and AI-generated images. It consists of over 100K AI images from five categories that cover three typical adversarial AI-art scenarios. 
\item We conduct a large-scale study to assess human users' ability to distinguish adversarial AI-art and present a detailed analysis of the results. 
\item We conduct the first large-scale benchmarking of the state-of-the-art open-source and commercial AI image detectors and show that most of them provide unsatisfactory detection performance. 

\end{enumerate}


\vspace{-2mm}
\noindent \textbf{Ethical Considerations.} This project aims to deepen the understanding of the security risks associated with AI-art. The ARIA dataset was collected via paid APIs provided by Midjourney, StabilityAI, StarryAI, and OpenAI, obeying their user policy. All of our AI-generated images will be shared with the research community. The copyright issue is discussed in Section \ref{subsec:copyright}. The user studies presented in the paper have been reviewed and approved by the Human Research Protection Program at the University of Kansas under STUDY00151343.

The rest of the paper is organized as follows: we present the background and preliminaries in Section \ref{sec:bg}. We introduce the ARIA dataset in Section \ref{sec:data}, followed by the user study and the benchmarking of open-source/commercial detectors in Sections \ref{sec:user} and \ref{sec:bench}, respectively. We discuss our findings in Section \ref{sec:dis}, survey the literature in Section \ref{sec:rel},  and conclude the paper in Section \ref{sec:con}.










\section{Background: AI-Generated Multimedia Content} \label{sec:bg}

\noindent \textbf{AI-Generated Content (AIGC).} AI-generated Content (AIGC) consists of digital content such as text, images, and music produced by advanced Generative AI (GAI) models instead of humans \cite{wu2023ai,liu2023check}. The history of GAI dates back to the 1950s with the creation of technologies like Hidden Markov Models \cite{rabiner1986introduction} and Gaussian Mixture Models \cite{reynolds2009gaussian}. The advent of transformer architecture \cite{vaswani2017attention} marked a significant milestone that led to a number of state-of-the-art models. For example, DALL-E-2 \cite{ramesh2022hierarchical} developed by OpenAI can quickly generate unique, high-quality images based on text descriptions. Examples in other domains include GPT-3 \cite{brown2020language}, Codex \cite{chen2021evaluating}, and Gopher \cite{rae2021scaling}.


\noindent \textbf{Diffusion Models in Image Generation.}
Diffusion models are a family of probabilistic generative models first introduced by \cite{sohl2015deep}. The mechanism is to progressively perturb the input by adding noises and then reverse this process to generate new samples \cite{yang2023diffusion}. Recent studies on diffusion models primarily focus on three approaches: denoising diffusion probabilistic models (DDPMs) \cite{ho2020denoising,nichol2021improved}, scored-based generative models (SGMs) \cite{song2020improved}, and stochastic differential equations (SDEs) \cite{song2021maximum}. These models have shown superior performance over generative adversarial networks (GANs) \cite{goodfellow2014generative} in image generation tasks \cite{dhariwal2021diffusion,ho2020denoising}. The success of diffusion models is largely due to their stable training processes and their foundation on likelihood estimates, which help overcome notorious issues in GANs like mode collapse \cite{croitoru2023diffusion}. Another notable example is SDXL \cite{podell2023sdxl}, which utilizes an extended latent space and more sophisticated reverse diffusion steps to enhance the quality and resolution of generated images. Also, the Diffusion Models can be applied conditionally, which makes it well-suitable for other computer vision tasks like super-resolution \cite{shang2024resdiff}, inpainting \cite{rombach2022high}, reconstruction \cite{takagi2023high}, and scene synthesis \cite{lei2023rgbd2}.


\noindent\textbf{Commercial Diffusion-based Generators.}
The commercial applications of diffusion models have revolutionized the way visual content is created. Notable platforms like Midjourney \cite{midjourney2024home}, DreamStudio \cite{dreamstudio2024}, StarryAI \cite{starryai2024home}, and OpenAI's DALL-E \cite{openai2024dalle} allow users to generate personalized images for a minimal cost, often less than five cents per image. These platforms typically offer two modes of creation: text-to-image, where users provide only text prompts, and image-to-image, which combines text prompts with existing images as seeds. These commercial services are significantly altering the landscape of fields like photography, fine art, and animation. 

\section{The \underline{A}dversa\underline{RI}al \underline{A}I-Art (ARIA) Dataset}\label{sec:data}

\subsection{Terminology: Real Images and AI-generated Images}\label{subsec:term}


In this paper, \textbf{Real Images} or \textbf{Human Images} are defined as images captured by conventional optical devices such as cameras and scanners or images created by human artists and then captured by optical scanning devices. \textbf{AI-generated Images} or \textbf{AI Images} are defined as images produced by generative models that only use text descriptions as prompts, i.e., all the visual contents are generated from text by AI models. 

We also consider two boundary cases. Graphic artworks, especially anime painted by human artists using painting/graphics software, are considered human-generated images, as they are protected by copyright in the same way as other creative works, e.g., original paintings and photographs. However, the detection of such images may pose different challenges than the detection of paintings/photographs. Meanwhile, images generated by AI models that take a \textit{seed image} and a \textit{text prompt} are considered AI images. We also examine how such images are different from images that are generated from pure textual prompts. 
\subsection{The Real Images and Annotation}\label{subsec:real}

In this project, our first objective is to collect a dataset of human-generated images and the corresponding (adversarial) AI images. This dataset will serve as a foundation for research on adversarial AI-arts and the detection of AIGC.

\vspace{2mm}\noindent\textbf{Topic Selection.} We identify three possible attack scenarios of adversarial AI-art, and further split these three scenarios into five detailed categories.

    \noindent$\bullet$~\textit{Social Media Fraud.} With the increased accessibility of AI image generators, we have witnessed an exponential growth of the number of AI images on social media, making it harder to distinguish reality from fake and enabling new forms of low-cost fraud \cite{deepfakedating,deepfakesocial}. For instance, the adversary could easily fake a celebrity lifestyle by posting AI-generated images of luxury goods, elite social events, exotic travel, and fine dining. Such fake profiles could be created and maintained at very low cost and employed in Internet scams. 
    We establish a specific dataset category named `\textit{Ins}', short for ‘Instagram-style images’, to represent lifestyle images that are typically shared on various social media platforms.
    
    \noindent$\bullet$~\textit{Fake News and Misinformation.} It has been widely reported that AI-generated content, including text and images, was adopted to produce disinformation and fake news articles \cite{wendling2024ai,chan2024one,Duffy_2024}. 
    For instance, a picture of a collision that involved a Tesla Cybertruck and a GMC Hummer EV gained popularity on the Internet in March 2024. It was later found to be AI-generated \cite{dintentdata2024ai}. 
    The concept of misinformation represents a very broad realm of topics. To better address this attack scenario, we define two categories: a `\textit{News}' category for general news images and a `\textit{Disaster}' category, which specifically captures images depicting emergencies that may trigger scare and panic.
    
    \noindent$\bullet$~\textit{Unauthorized Art Style Imitation.} Generative models have the capacity to mimic the style of renowned artists and produce forged artwork. This will potentially produce fraud and violate consumer rights. 
    Also, AI generators can easily replicate famous anime characters' images by referring to the name, e.g., Edward Elric or Hatsune Miku, in the prompts. If the generated content is used without proper licensing or authorization, these AI-generated images violate the copyright held by the entities who own these characters \cite{Thompson_2024}. 
    We define two different AI image categories to represent the legally risky production of these two popular art types. First, the `\textit{Art}' category contains the classic fine art images. The second category, `\textit{Pixiv}', is named after one of the most popular platforms for artists to share copyright-protected and original illustrations. This platform is losing users to AI-generated art \cite{wei2024understanding}. Therefore, `Pixiv' represents another potential area of research on the social and legal impacts of AI-arts.

\noindent\textbf{Collection of Real (Human-generated) Images.}
To select human image datasets for the five categories defined above, we followed two criteria: (1) we expect the datasets to be diverse, representative, and contain a reasonable amount of images; (2) to ensure that all the images were human-generated, we select the datasets that were collected before 2022, i.e., before generative AI was able to produce high-quality images that appeared like human images. 

We selected dataset \textit{InstaNY100K} \cite{instany100k} for `Ins'. This dataset, last updated in Dec 2021, includes 10,000 images from real Instagram posts. For `News', we chose the \textit{N24News} dataset \cite{n24news,wang2022n24news} extracted from the New York Times from 2010 to 2020. For `Dis', we chose the \textit{Disaster Dataset} \cite{disaster2012dataset}, which contains photos of various disaster scenes including damaged infrastructure, fire, human injury, etc. Its last update was in 2022. For `Art', we used dataset \textit{Best Artworks of All Time} \cite{art2023dataset}, a collection of artworks of the 50 most influential artists, updated in 2019. Lastly, we selected \textit{Pixiv Top Daily Illustration 2018} \cite{pixiv2019dataset} for `Pixiv', which contains 68,800 popular Pixiv images in 2018. 
To minimize the risk of including AI images, the dataset owner filtered the illustrations with the tag `Original'.


To ensure a reasonably sized dataset and equal representation from all categories, and to reduce the cost, we randomly sampled approximately 3,000 images from each category except for `Art'. We selected 50,000 images for `Art' because this dataset is more diverse than the others. Eventually, the ARIA dataset contains 17,129 real-world images with a total file size of 1.4GB.


\noindent\textbf{Annotation.} 
We intend to pair AI images with human images. For instance, when we have a human image of Claude Monet's \textit{Water Lilies}, we would like to prompt each AI generator to produce an image for ``Claude Monet's Water Lilies''. 
To achieve this, we designed a systematic process to annotate the collected real images with detailed textual descriptions, which are subsequently used in the prompts for the AI image generators. 

\noindent\textit{1. Automated Image Description.} First, we utilized MidJourney's ``describe'' function \cite{midjourney2024home} to generate four comprehensive text annotations by identifying the key visual elements within each image. 

\noindent\textit{2. Text Prompt Synthesis and Optimization.} We observed issues with MidJourney's ``describe'' function--it frequently generates and includes meaningless hashtags, non-word strings, and random names in the description. The four descriptions for the same image are also moderately inconsistent. We further invoke ChatGPT to correct the mistakes described above and integrate the four different descriptions to create a comprehensive text prompt. 

    
\noindent\textit{3. Keyword Enhancement of Text Prompts.} To ensure the text prompts reflect the specific characteristics of each attack scenario, we further add category-specific keywords to the beginning of the annotation. For example, in the `Art' category, we added keywords for each image to specify this is an artwork by the artist (e.g.,  Claude Monet), whose names were extracted from the dataset \cite{art2023dataset}. In the `Dis' category, terms such as `photography', `disaster', and `incident' are added. 


\subsection{AI-Art Generation}

In this section, we detail the design of the prompts and the processes to employ different platforms to generate AI-arts. 

\noindent\textbf{Commercial AI-Art Generators.} 
Given the rapid advancement of AI image generators, the threshold of employing AI-generated fake images in fraud/scams has significantly diminished. Various AI image generation platforms enable attackers to effortlessly generate realistic images. 
In this study, we selected four widely used AI generators: Midjourney \cite{midjourney2024home}, DreamStudio \cite{dreamstudio2024}, StarryAI \cite{starryai2024home}, and DALL-E \cite{openai2024dalle}. They are popular for reliability, accessibility, and relatively low cost, making them ideal candidates for our image generation task.

\noindent\textbf{Mode of Generation and Prompt.} To maintain consistency, we use uniform prompt structures for all generations. Our study employs two different generation modes, each requiring its specialized prompt.

\noindent$\bullet$~\textit{Text-to-Image (T2I) Generation.} In the T2I mode, images are created from scratch by feeding solely the text descriptions to the generators. We utilized the text annotations gathered during the annotation phase as prompts.

\noindent$\bullet$~\textit{(Image+Text)-to-Image (IT2I) Generation.} Images are generated from two pieces of seed information: (1) the text descriptions obtained from the annotation process and (2) the corresponding real image. This method combines the visual cues from the seed image with textual information to generate new images.

\noindent\textbf{Parameters and Settings.} 
The common parameters for all the platforms are: (1) \textit{Aspect Ratio:} For all platforms, we used the default aspect ratio of 1:1. (2) \textit{Resolution:} The default resolution varies on different platforms. Midjourney and DALL-E are set to 1024x1024 pixels, and Dream Studio is 516x516 pixels. Starry AI only provides a True/False selection for \textit{High Resolution}, for better image quality, we set it to `True'. (3) \textit{Strength:} This parameter controls the respective influences of image and text seeds in the IT2I mode. We set it to be 50\% - 50\% to ensure that the text description and seed image have equal impacts.

Besides these common parameters, three platforms have specific features on generation models and processes.



\noindent$\bullet$ \textit{Midjourney's} Discord bot produces four output images in each generation. We collect the first image among these four. We employed model 5.2, which was the newest model during our data collection\cite{midjourney2024home}, for image generation.




\noindent$\bullet$ \textit{Dreamstudio's} API uses the newest version of the Stable Diffusion model. Two specific parameters are worth noting: (1) \textit{Sampler} determines how to extract the final image from the latent space. We use the default value SAMPLER\_K\_DPMPP\_2M. (2) \textit{cfg\_scale (Conditional Free Guidance Scale)} controls the compliance of text descriptions. Since our text annotations accurately represent source image content, we choose CFG\_scale value 8.0 for text-to-image generation and the default value of ``7.0'' for image-to-images generation.

\noindent$\bullet$  \textit{Starry AI's} API invokes various models for specific generation tasks. We use \textit{Photography} for the `Disaster', `Instagram', and `News' categories. We select model \textit{Anime} for `Pixiv', model \textit{Argo2} for the T2I mode of `Art', and model \textit{Argo} for IT2I for `Art' (Starry AI disabled \textit{Argo2} during our generation).

\noindent$\bullet$ \textit{DALL-E 3} was used for text-to-image generation. However, since DALL-E 3 does not provide image-to-image functionality, we use DALL-E 2 for IT2I. DALL-E has a parameter `quality' that controls the level of details of the generated image, we set it to the default value ``standard''.

\subsection{The ARIA Dataset}


\begin{figure}[t]
\centering
	\includegraphics[width=0.98\linewidth]{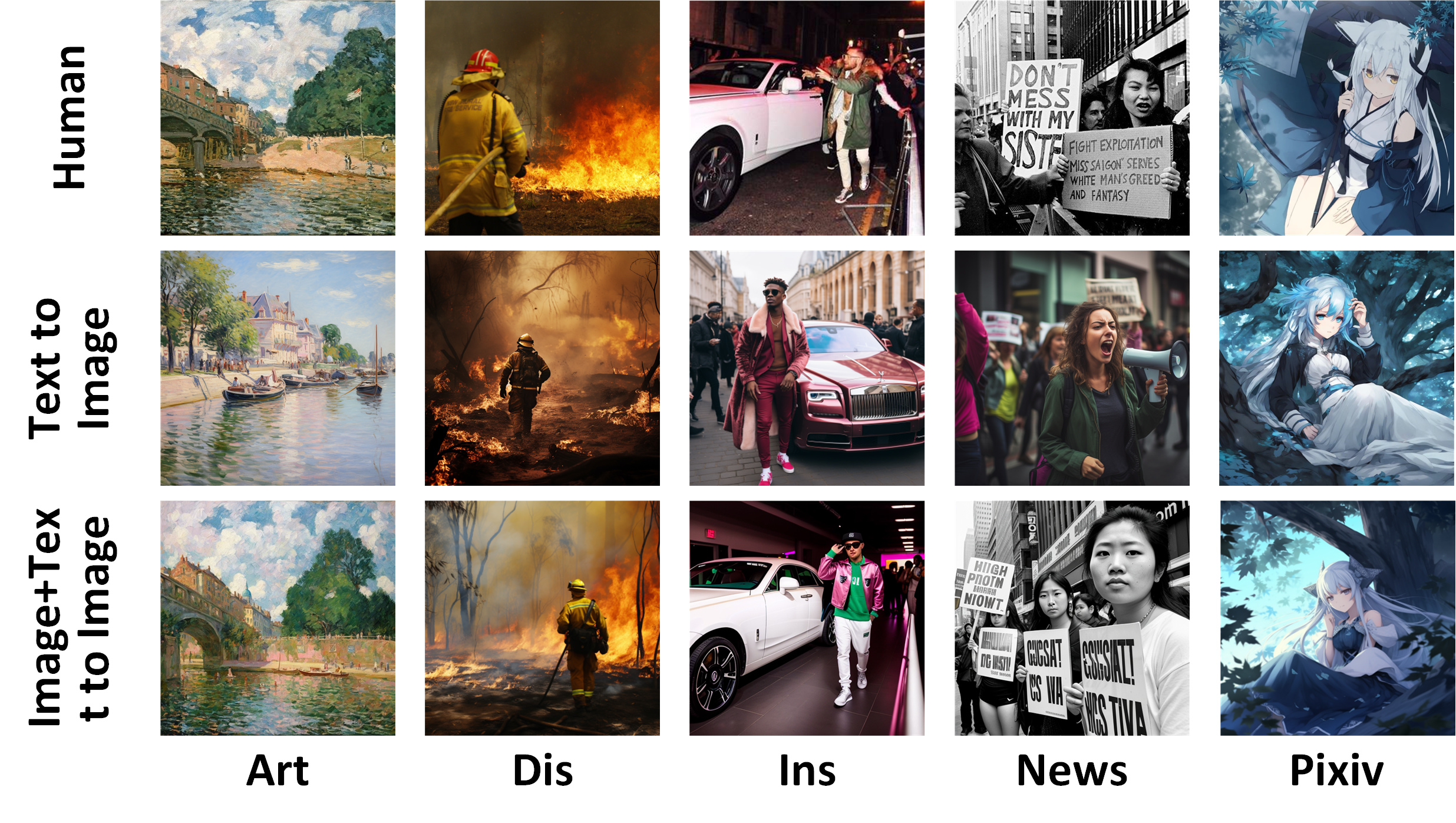}\vspace{-5mm}
	\caption{Sample images from the ARIA dataset. \vspace{-1mm}}
	\label{fig:exp}
\end{figure}

The final ARIA dataset contains 127,046 AI-generated images and 17,129 real images, as detailed in Table \ref{tbl:generation}. Despite using identical prompts and real images across platforms, the quantity of images produced by each generator differed. 

\begin{table}[t]
\centering
\caption{The number of images by category.}\label{tbl:generation}\vspace{-1mm}
\setlength{\tabcolsep}{3pt}
\renewcommand{\arraystretch}{1.2}
\begin{tabular}{c|cccc|cccc|c|c}

\hline
\multirow{2}{*}{Category} & \multicolumn{4}{c|}{Text-to-Image} & \multicolumn{4}{c|}{Image-to-Image} & \multirow{2}{*}{Real} & \multirow{2}{*}{\textbf{Total}} \\\cline{2-9}
         & MJ & DS & SA & DA & MJ & DS & SA & DA &  & \\\hline
Art      & 4999 & 5171 & 2160 & 4925 & 4939 & 5186 & 5180 & 5188 & 5327 & 43075 \\
Dis      & 2762 & 2790 & 2831 & 2896 & 2479 & 2790 & 2830 & 2947 & 2963 & 25288 \\
News     & 3027 & 2866 & 2849 & 2974 & 3001 & 2764 & 3017 & 3029 & 3032 & 26559 \\
Pixiv    & 2400 & 2642 & 2780 & 2355 & 2045 & 2636 & 2808 & 2508 & 2814 & 22988 \\
Ins      & 2961 & 2809 & 2895 & 2960 & 2907 & 2763 & 2984 & 2993 & 2993 & 26265 \\
\textbf{Total} & 16149 & 16278 & 13515 & 16110 & 15371 & 16139 & 16819 & 16665 & 17129 & \textbf{144175} \\
\hline
\end{tabular}\vspace{-2mm}
\end{table}

The generators also deployed content filtering mechanisms. For instance, \textit{Midjourney} blocks the text inputs that contain adult and gore content \cite{midjourney2024home}. Although not explicitly mentioned on its official website, \textit{Midjourney} also automatically stops generation once it detects potentially inappropriate content in the seed image. Such censorship for image prompts is even stricter than for text prompts. For example, images involving human body nudity in the disaster and artwork datasets would stop the generation immediately. Hence, we had fewer IT2I images than T2I images. Moreover, the censorship mechanism also produces false positives, where harmless images are flagged.
Dream Studio chose to blur out the generated images when inappropriate or offensive content is found
\cite{dreamstudio2024}. We deleted the blurred images from the dataset. Furthermore, content filters from different generators are often triggered by different prompts, and the resulting image count for each generator is different. 

For Starry AI, 
the \textit{Argo2} model for the `Art' image generation is disabled from Starry AI API during our generation. To maintain the consistency of our generated images, we opted to pause the generation instead of switching to \textit{Argo}. We plan to re-generate the remaining images once \textit{Argo2} is back on the market. 


The total cost for all AI-generated images is approximately \$3,550, including \$810 for \textit{Midjourney}, \$120 for \textit{Dream Studio}, \$1,870 for StarryAI, and \$1,200 for DALL-E. The total file size for all the generated images exceeds 100 GB.  Notably, DALL-E returns images with the highest JPEG quality, so that the size of each 1024$\times$1024 image could be 3MB or higher. 

\subsection{Copyright and License} \label{subsec:copyright}

We acknowledge the ongoing challenges regarding copyright issues of AI-generated images. We would like to clarify the copyright issues of the set of images we used to create new AI-generated images and the set of images created by us using the original images with AI tools. 

For the real image datasets, the licenses differ by dataset. The `Art' dataset \cite{art2023dataset} is shared under a CC-BY-NC-SA 4.0 license, allowing ``[a]dapt — remix, transform, and build upon the material''. The owner of the `Ins' dataset \cite{insta2022dataset} labeled the dataset as public domain under CC0 1.0 Deed, which ``dedicated the work to the public domain by waiving all of his or her rights to the work worldwide under copyright law ...''. The `Dis' dataset does not come with any license. The `Pixiv' and `News' datasets both contain copyrighted images. We use these three datasets under the Fair Use clause for teaching, scholarship, and research (Section 107 of the Copyright Act). We provide links to these datasets instead of re-sharing them in the ARIA dataset. 

For AI-generated images, all four platforms' policies explicitly declare that the creator of the images (i.e., the authors of this paper) owns them. For instance, Midjourney claims that ``subscribers own all the images they’ve created, even if their subscription has expired, and they’re free to use those images however they’d like'' \cite{midj2024copyright}. 
DALL-E's Content Policy states that ``[s]ubject to the Content Policy and Terms, you own the images you create with DALL·E, including the right to reprint, sell, and merchandise – regardless of whether an image was generated through a free or paid credit'' \cite{dalle2024copyright}. 
Based on the content policies stated by each tool respectively, we, therefore, assert the ownership of all AI-generated images in the ARIA dataset. We share them under the CC-BY-NC-SA license.

\section{The User Study} \label{sec:user}


\subsection{The Design}

\begin{figure}[t]
\centering
	\includegraphics[width=0.98\linewidth]{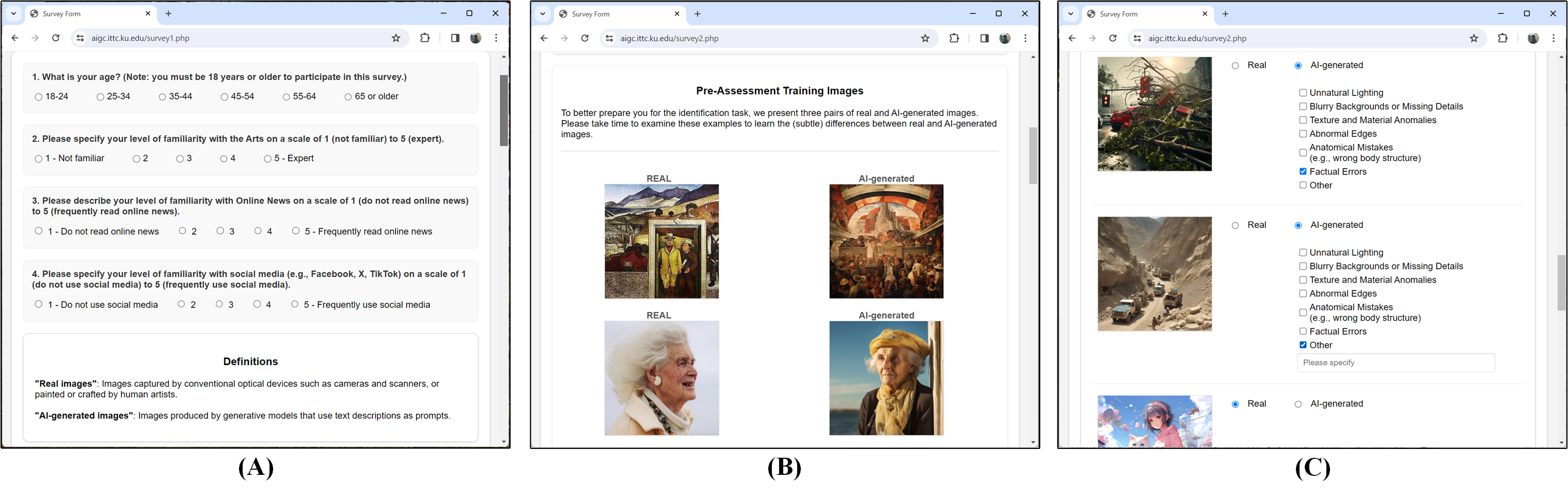}\vspace{-2mm}
	\caption{The user study: (A) The collection of users' background information; (B) The ``training'' samples for users with references ($U_R$); (C) The main survey.\vspace{-2mm}}
	\label{fig:survey}
\end{figure}

In this section, we present our user study that aims to learn if  Internet users can identify AI-generated (adversarial) images. In particular, we aim to answer the following research questions: \textit{RQ1.} Could real-world users, with or without reference images, distinguish between real images and AI-generated images? \textit{RQ2.} What visual clues do they use in identifying AI-generated images? \textit{RQ3.} Does the users' capability of identifying AI images vary for different image topics? \textit{RQ4.} Do popular AI generators produce images that differ in identifiability? \textit{RQ5.} Does the user's background knowledge, e.g., familiarity with the art, contribute to their capability of identifying AI-generated images?

Based on the users' knowledge of AI-generated images, we present two detection models: (I) \textbf{Referenceless Users ($U_0$).} This represents the majority of Internet users, who have not been explicitly exposed to AI-generated images, especially in a side-by-side comparison with real images. (II) \textbf{Users with References ($U_R$).} With the growing popularity of AIGC, especially with the recent media coverage, some users may be aware of AI-generated images. In this user study, we mimic this type of user by providing the surveyee with 3 random pairs of real and AI-generated images in a side-by-side setting and asking them to examine the sample images before continuing to the questionnaire.  

In this user study, an IRB information statement is first provided to the user, followed by the link to the questionnaire. In the questionnaire, we first ask four questions about the user's background (Figure \ref{fig:survey} (A)): (1) the user's age range, e.g., 18-24, 25-34. (2) The user's familiarity with the Arts on a scale of 1 (not familiar) to 5 (expert). (3) The user's familiarity with Online News on a scale of 1 (do not read online news) to 5 (frequently read online news). And (4) the user's familiarity with social media (e.g., Facebook, X, TikTok) on a scale of 1 (do not use social media) to 5 (frequently use social media). In data analysis, we will use such information to answer research question RQ5.  

Next, we present the definitions of AI-generated images and real images (Figure \ref{fig:survey} (A)). Approximately 50\% of the surveyees will be randomly assigned to the $U_R$ scenario, where three pairs of randomly selected real and AI-generated images are displayed with a short text instruction, as shown in Figure \ref{fig:survey} (B). 

Ten images, randomly selected from different categories of AI-generated images and real images are then displayed to the user. As shown in Figure \ref{fig:survey} (C), the surveyee is asked to identify each image as ``real'' or ``AI-generated''. If the user marks an image as ``AI-generated'', checkboxes will show up to ask her to provide the selection rationale. As there does not exist any paper in the community that discusses how AI-generated images could be visually identified, we referred to several news articles to summarize a list of possible clues \cite{Writer_2024,Kavafian_2023,Steele_2024}. Finally, the user could select ``other'' and enter her own rationale. 

\subsection{The Results and Analysis}

We recruited volunteers (faculty, staff, and students) from the University of Kansas as well as our collaborator institutions to respond to the survey. In three weeks, we have received 4720 annotations from 472 participants. Here we present our statistics of the surveys and answer the research questions. 

\noindent$\bullet$~\textit{RQ1. Could real-world users, with or without reference images, distinguish between real images and AI-generated images?} 

The average accuracy for referenceless users $U_0$ was 65.24\%. In particular, 79.87\% of the real images were correctly identified, while only 61.58\% of the AI-generated images were correctly identified. Users with reference images ($U_R$) performed slightly better, with an average accuracy of 68.00\%, an accuracy of 79.76\% for real images, and an accuracy of 65.06\% for AI-generated images. From the results, we can observe the following: (1) it is highly challenging for human users to identify AI-generated images, with or without references; (2) Human users are significantly more likely to mistakenly identify AI-generated images as real (38.42\% for $U_0$ and 34.94\% for $U_R$) than to mistakenly identify real images as AI-generated (20.13\% for $U_0$ and 20.24\% for $U_R$). 

\noindent$\bullet$~\textit{RQ2. What visual clues do they use in identifying AI-generated images?}

When a user labels an image as AI-generated, he/she is asked to select a reason from a provided list or provide her/his own reason. For Scenario 1 ($U_0$), 378 images were annotated as ``Texture and Material Anomalies'', which was the most popular reason provided by the users. 350 out of 378 (92.6\%) images were correctly identified as AI-generated. Meanwhile, 337 images were annotated as ``Anatomical Errors'', the second most popular reason. 316 out of 337 (93.8\%) images were correctly identified. All other selections were significantly less popular. Another notable observation is that Midjourney and DALL-E generate significantly fewer images with anatomical errors (11\% and 10.1\% of all correctly identified AI-generated images) than the other two generators (22\% and 21\% of all identified AI-generated images). Our further examination of the dataset also confirmed that the Midjourney/DALL-E-generated images are significantly less likely to show errors like extra fingers or distorted faces. 

2.31\% of $U_0$ and 1.26\% of $U_R$ provided their own reasons besides the provided list. They were examined and coded by two graduate students. The provided rationales were highly consistent across $U+0$ and $U_R$.  The top 3 reasons are: (1) detailed explanations of the factual errors, such as ``illegible text''. They account for 48.9\% of user-provided rationales for $U_0$ and 56.2\% for $U_R$. (2) Subjective feelings (36.7\% for $U_0$ and 22.9\% for $U_R$), such as ``feels ai,'' ``art style looks AI.'' And (3) detailed explanations of anatomical mistakes (10.2\% for $U_0$ and 18.8\% for $U_R$), such as ``hands look out of proportion,'' ``girl's collarbone.'' 


\begin{table}[t]
\centering
\caption{The detection accuracy (\%) of real and AI-generated images by human users.}\label{tbl:user}\vspace{-2mm}
\setlength{\tabcolsep}{0.2em}
\begin{tabular}{c|ccccc|c||ccccc|c}

    & \multicolumn{6}{|c||}{Referenceless Users ($U_0$)} & \multicolumn{6}{c}{Users with References ($U_R$)} \\\cline{2-13}
 	& art &	disaster & news & pixiv & ins & Avg & art &	disaster & news & pixiv & ins & Avg \\\hline
Dream Studio & 	55.4& 	61.2& 	78.6& 	66.3& 	75.6& 	67.4 & 58.3 & 	70.6 & 	69.7 & 	70.5 & 	67.5 & 	67.3 \\
Midjourney & 	38.1& 	48.4& 	62.8& 	51.5& 	64.4& 	52.5 & 50.4 & 	69.7 & 	66.3 & 	60.6 & 	65.2 & 	61.2 \\
Starry AI & 	47.7& 	60.6& 	70.8& 	55.4& 	64.6 & 	60.0 & 38.5 & 	64.4 & 	76.4 & 	63.8 & 	64.8 & 	60.7 \\
DALL-E & 68.9& 	94.3& 	89.1& 	61.5& 	93.8& 	80.7 & 61.5 & 	78.9 & 	92.0 & 	67.9 & 	87.5 & 	77.9 \\
\hline
Avg. & 	50.4 & 	59.2 & 	71.7 & 	58.9 & 	68.7 & 	61.6 & 50.5 & 	67.8 & 	72.5 & 	66.8 & 	69.8 & 	65.1 \\\hline\hline
real & 	81.2 & 	80.4 & 	86.0 & 	71.7 & 	81.5 & 	79.9 & 85.0 & 	76.7 & 	81.6 & 	77.9 & 	77.6 & 	79.8 \\
\end{tabular}\vspace{-5mm}
\end{table}

\noindent$\bullet$~\textit{RQ3. Does the users' capability of identifying AI-generated images vary for different image topics, e.g., news images or artwork?}

A breakdown of the users' identification accuracy for images generated by each platform and images of each topic is shown in Table \ref{tbl:user}. We have the following observations: (1) In both scenarios ($U_0$ and $U_R$), users are least capable of identifying art images. This could be partially explained by the fact that our survey participants are generally less familiar with the arts (average familiarity: 2.56/5) than online news (average familiarity: 3.39) and social media (average familiarity: 3.60). (2) Users appear to be better at identifying AI-generated images in news. This may be partially explained by the fact that news images in ARIA were more likely to be taken closer to the scene with more details and human characters that are more identifiable to the evaluators. 

\noindent$\bullet$~\textit{RQ4. Do popular AI generators produce images that differ in identifiability?}

Images generated by DALL-E appear to be significantly more identifiable to human users than the other generators, with Dream Studio being a distant second. Our examination of the data also shows that DALL-E-generated images have unique lighting and texture features that make them easily identifiable. 

\noindent$\bullet$~\textit{RQ5. Does the user's background knowledge, e.g., familiarity with the art or social media, contribute to their capability of identifying AI-generated images? }

Table \ref{tbl:exptise} presents a breakdown of the users' self-claimed domain expertise and their identification accuracy on the relevant topics. We only observed a strong correlation between the identification accuracy in `Art' and the self-claimed familiarity with art: $r_{pearson}=0.945$ for $U_0$. However,  $r_{pearson}=0.5$ decreased to 0.483 for $U_R$, which could be because some reference samples may compensate for low familiarity. In most cases, the self-claimed domain expertise (or the surveyee's age) does not appear to show a significant correlation with their capability to identify AI images. That is, even for users who are highly familiar with online news and online social networks, it is still challenging to recognize AI-generated adversarial images on the Internet. The attack scenarios we presented in Section \ref{subsec:real} pose a high risk to real-world users. 

\begin{table}[t]
\centering
\caption{The impact of self-claimed domain expertise on detection accuracy (\%) of real and AI-generated images. AF: familiarity with art; NF: familiarity with online news; SMF: familiarity with social media; ---: insufficient data for statistical significance. }\label{tbl:exptise}
\vspace{-2mm}
\setlength{\tabcolsep}{0.3em}
\begin{tabular}{c|ccccc|c||ccccc|c}

    & \multicolumn{6}{|c||}{Referenceless Users ($U_0$)} & \multicolumn{6}{c}{Users with References ($U_R$)} \\\cline{2-13}
 	expertise & 1 & 2 & 3 & 4 & 5 & Avg & 1 & 2 & 3 & 4 & 5 & Avg \\\hline
art vs. AF & 39.5 & 	41.4 & 	59.3 & 	63.0  & --- & 	50.4 & 48.4 & 	40.0 & 	55.3 & 	58.5 & 	50.0 & 	50.5 \\ 
news vs. NF & --- & 75.0 & 	82.3 & 	64.4 & 	67.6 & 	71.7 & 78.6 & 	65.3 & 	79.2 & 	70.3 & 	70.0 & 	72.5 \\
disaster vs. NF & 	--- & 	55.1 & 	62.7 & 	50.0 & 	67.6 & 	56.7 & 53.1 & 	66.7 & 	69.1 & 	65.5 & 	81.0 & 	67.8\\
ins vs. SMF & --- & 61.0 & 	75.0 & 	65.9 & 	70.5 & 	67.9 & 71.4 & 	73.0 & 	66.7 & 	60.0 & 	68.7 & 	66.7 \\
disaster vs. SMF &	--- & 	54.8 & 	57.1 & 	49.6 & 	68.8 & 	59.2 & 50.0 & 	52.3 & 	65.8 & 	62.5 & 	81.7 & 	67.8 \\ 
\end{tabular}\vspace{-4mm}
\end{table}

\vspace{2mm}\noindent\textbf{Population Bias.} We acknowledge that our population may not accurately represent the majority of Internet users. It is likely that our surveyee population is more educated and has higher familiarity with online news and online social networks. However, our correlation study between detection accuracy and user background shows that such population bias is unlikely to affect the key findings of this study since the users' expertise is not a contributing factor to their capability of identifying AI-generated images in most scenarios. 

\section{Benchmarking AI-image detectors}\label{sec:bench}


In this section, we assess the performance of state-of-the-art AI image detectors.

\begin{table}[t]
    \caption{Accuracy (\%) of commercial and open-source detectors on the ARIA dataset. Bold: best performance for each category (open/commercial) and each generator. }\label{tbl:benchmark}\vspace{-2mm}
    \setlength{\tabcolsep}{0.15em}
    \centering{
    \begin{tabular}{c|c|cccc|cccc}
    \hline
    & \multirow{2}{*}{HUM} & \multicolumn{4}{c|}{T2I} & \multicolumn{4}{c}{IT2I} \\\cline{3-10}
    & & MJ & DS & SA & DA & MJ & DS & SA & DA \\\hline
    \multicolumn{10}{c}{Open-source Detectors}  \\    \hline  
    Organika-ViT \cite{organika2024sdxl} &  72.83 & \textbf{90.92} & \textbf{98.31} & 55.48 & 77.17 & \textbf{77.32} & \textbf{54.19}  & 63.90 & 26.27 \\ 
    umm-maybe-ViT \cite{maybe2024detector} &  81.01 & 22.59 & 24.22& 39.67 & 71.45 & 25.27 & 20.23  & 40.18 & 31.62 \\
    Nahrawy-swin \cite{Nahrawy2023aiornot} & 73.93 & 67.72 & 72.24 & \textbf{89.89} & \textbf{90.71} & 74.75 & 31.09  & \textbf{88.18} & \textbf{68.62} \\ 
    Wvolf-CNN \cite{wvolf2024cnn} & 62.92 & 3.22 & 3.73 & 3.85 & 2.66 & 3.69 & 5.00 & 4.58 & 8.56 \\ 
    Nodown-stylegan2 \cite{gragnaniello2021gan} & \textbf{100} & 0.19 & 0.03 & 0.15 & 0.32 & 3.09 & 0.27 & 0.49 & 8.88 \\
    Nodown-progan \cite{gragnaniello2021gan} & 99.82 & 23.56 & 1.18 & 11.60 & 0.00 & 28.10 & 2.12 & 11.35 & 36.24 \\
    Spectrum-pixel \cite{he2021beyond} & 98.91 & 2.23 & 12.34 & 31.26& 5.68  & 3.34 & 7.60 & 0.88 & 0.48 \\
    Spectrum-stage5 \cite{he2021beyond} & 98.58 & 5.75 & 15.40 & 2.09 & 11.44 & 5.63 & 20.16 & 16.54 & 29.76 \\
    CNNSpot \cite{wang2020cnn} & 99.44 & 0.43 & 1.12 & 0.12 & 0.16 & 0.65 & 2.21 & 0.11 & 2.80 \\
    
    \hline\multicolumn{10}{c}{Commercial Detectors}  \\    \hline       
    Illuminarty \cite{illuminarty} &   92.72 & 66.45 & 80.52 & 81.49 & 90.47 & 1.42 & 31.09  & 61.29 & 65.01 \\
    sightengine \cite{sightengine} &  \textbf{98.72} & \textbf{96.40} & \textbf{95.84} & \textbf{99.04} & \textbf{100} & \textbf{95.76} & 25.12  & \textbf{93.40} & 43.76 \\
    Is it AI? \cite{isitai} & 77.20 &  74.10 & 72.65 & 77.50 & 77.62 & 61.63 & 32.49 & 65.40 & 32.57   \\
    Content at Scale \cite{contentatscale} &  75.03 & 24.13 & 24.87 & 40.17 & 71.28 & 25.90 & 23.83  & 41.10 & 50.40 \\
    Fake Image Detector \cite{fakeimgdetector} &  35.57 & 61.41 & 49.87 & 47.28 & 76.22 & 86.74 & \textbf{60.87}  & 61.36 & \textbf{68.03} \\

    \hline\end{tabular}}\vspace{-3mm}
\end{table}

\noindent\textbf{Commercial Detection Services.}
Due to the widespread concerns about the misuse of AI-generated images, there has been a strong demand for professional services that can distinguish between human-generated and AI-generated images. Multiple service providers are active in this market. 
They claim high detection accuracy and have been employed across several industries, including social media and review sites. Most of these detectors provide user-friendly web interfaces for general users. In this work, we choose the detectors that provide APIs for professional users: Illuminarty \cite{illuminarty}, Sight Engine \cite{sightengine}, Is it AI? \cite{isitai}, Content at Scale \cite{contentatscale}, and Fake Image Detector \cite{fakeimgdetector}.

\noindent\textbf{Open-source Detectors.}
We choose the following open-source detectors that are available on GitHub or Hugging Face: SDXL \cite{organika2024sdxl} (Organika-ViT), Umm-maybe \cite{maybe2024detector} (umm-maybe-ViT), AI-or-not \cite{Nahrawy2023aiornot} (Nahrawy-swin), Deepfake based CNN Detector by Rudolf Enyimba \cite{wvolf2024cnn} (Wvolf-CNN), GAN based detector by Gragnaniello et al. \cite{gragnaniello2021gan} (Nodown-stylegan2, Nodown-progan), Beyond the Spectrum detector by He et al. \cite{he2021beyond} (BeyondtheSpectrum-pixel, BeyondtheSpectrum-stage5), CNNSpot detector by Wang et al.\cite{wang2020cnn} (CNNSpot). 

\noindent\textbf{Experiments.} We evaluate the models on the ARIA dataset on an RTX4060Ti GPU. For the commercial detectors, we invoke their APIs with the default settings. For the open-source detectors, we use the pre-trained models shared by the authors on GitHub or Huggin Face. Open-source detectors usually return two values for each image: a label of ``human'' or ``artificial'' and a confidence level in [0, 1]. 
Most detectors return confidence levels of 95\% or higher for more than 80\% of the testing images, while the confidence drops to below 75\% for fewer than 8\% of images. We evaluate the detectors with all the human-generated and AI-generated images in the ARIA dataset. The results are shown in Table \ref{tbl:benchmark}. 
\noindent\textbf{Results and Analysis.}
From the results reported in Table \ref{tbl:benchmark}, we have the following observations: (1) The open-source detectors provide unsatisfactory detection accuracy, i.e., mostly below 70\% in detecting AI images. For reference, an accuracy of 50\% on this binary classification task is equivalent to random guesses. (2) For open-source detectors, there is a clear trade-off in the detection accuracy of human images and AI images. That is, a higher accuracy for human images almost always leads to a lower accuracy for AI images. (3) Most of the detectors also have a strong tendency to label most of the images as human-generated, indicating a potential bias and limitation of current detection tools. (4) Some commercial detectors provide better performance and a better balance between the accuracy of human and AI images. (5) Most of the commercial detectors perform worse on IT2I images. This could be explained by the fact that seed images were used in generating the IT2I images, which resulted in higher similarity between human and IT2I images. In particular, some generators, such as Dream Studio, produced IT2I images that appear almost identical to the seed images, despite a 50\%-50\% image-text ratio being set. 

\begin{figure}[!t]
\centering\vspace{-2mm}
\includegraphics[width=0.7\columnwidth]{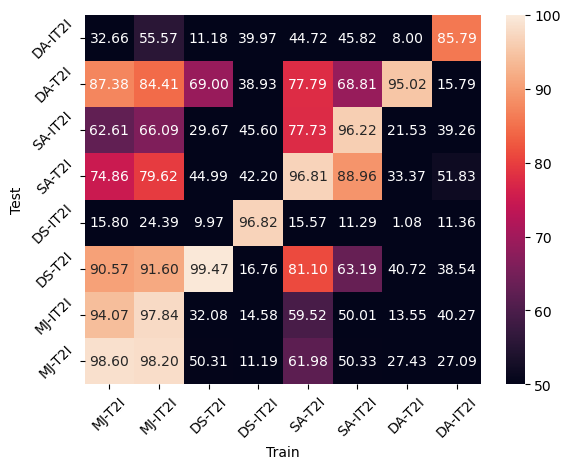}\vspace{-5mm}
\caption{F1-score (\%) of ResNet-50 trained and evaluated on different subsets.}
\label{fig:resnet}\vspace{-5mm}
\end{figure}

We further adopted a ResNet-50 model \cite{he2016deep} for AI image identification. We trained and tested the model on ARIA data from each generator and each generation mode (T2I or IT2I). For each subset, we allocated 70\% of data for training and 30\% for testing, ensuring no overlap between the two. Additionally, we maintained the same testing data throughout our cross-validation to preserve consistency and reliability in our results. The results of the 8$\times$8 cross-validation are shown in Figure \ref{fig:resnet}. The model gives the highest F1 scores when it is trained and tested with the same generator+mode combination. However, the F1-score decreases when the testing data deviates from the training data, even by just switching the generation mode with the same generator. Upon reviewing our testing results, we observed that this decline is more pronounced in AI-generated images, while the accuracy for real images remains consistently high across all tests, due to all models being trained on the same human images. This also might be another reason why the commercial and open-source detectors perform better on real images beyond bias, as there are many real image datasets available for training, but datasets for AI-generated images are not comprehensive enough, highlighting a gap in detector training resources. 



We do not compare our classifier with the open-source/commercial detectors. Such comparisons would not be fair since they are not trained on ARIA data, while the commercial detectors use complex models trained on larger datasets.

\section{Discussions}\label{sec:dis}

\noindent\textbf{Security Analysis of the Adversarial AI Art.}
In a practical attack, the adversary may generate images that match the attack scenario by simply providing suitable prompts. Such generation is particularly effortless and effective, especially since attackers have the advantage of selecting the most convincing images from many generated choices. Our user study also revealed that identifying these AI-generated images by human eyes is highly challenging, with an average accuracy below 65\%. The detection performance also depends on various factors like the image content, generation methods, and the generator. Although tools like Sight Engine appear to be effective in detecting AI images, most other detectors available to the general public fall short. Furthermore, in real-world scenarios, detectors are typically used only when the image already triggers suspicion, which appears to be unlikely for most of the AI images. In summary, adversarial AI-art poses real challenges to the community, while highly effective and practical solutions are still on the way. Moreover, since IT2I generations are relatively underexplored in the literature, they represent a potential area for future research. Finally, we also hope to stimulate awareness among the users who may fall victim to such attacks.

\noindent\textbf{Limitations.} With the rapid evolution of AI models, our dataset might become less representative over time. However, the performance issues in state-of-the-art detectors suggest the continued relevance of our findings. Budget constraints limited our use of more expensive detectors like AIorNot. Although we strive to ensure our real image dataset's authenticity, the possibility of encountering edited (photoshopped) images cannot be entirely ruled out. However, the distinction between edited images and AI-generated images does not fundamentally impact our research goals, as they represent different types of data manipulation.

\section{Related Work}\label{sec:rel}
\noindent\textbf{AI-generated Image Dataset.} The majority of existing works on AI-generated images concentrated on GAN-based deepfakes \cite{rossler2019faceforensics++,wang2020cnn}. Recent studies pay attention to diffusion-based models. Fake2M \cite{lu2024seeing}, DE-FAKE \cite{sha2023fake}, DiffusionForensics \cite{wang2023dire}, CiFAKE \cite{bird2024cifake}, ArtiFact \cite{rahman2023artifact} and GenImage \cite{zhu2024genimage} are based on conventional real-life datasets like scenes (LSUN \cite{yu2015lsun}), objects (MSCOCO \cite{lin2014microsoft}, CIFAR10 \cite{krizhevsky2009learning},  ImageNet \cite{deng2009imagenet}), or faces (CelebA-HQ \cite{karras2018progressive}). WildFake \cite{hong2024wildfake} and DiffusionDB \cite{wang2023diffusiondb} include fake images sourced from open-source websites or servers. All these works serve general purposes. DDDB \cite{wang2023benchmarking} and \cite{ha2024organic} studies AI-generated artworks. Our dataset is unique in three ways: (1) We employ a systematic approach for prompt design to enhance the association between human-generated and AI-generated images. In contrast, many datasets use simple prompts like ``photo of {label}" \cite{zhu2024genimage,wang2023dire}, leading to simple and heterogeneous image content. (2) ARIA focuses on the adversarial applications of AI images that pose practical risks in the real world. (3) The existing datasets mainly focus on T2I images, and ours incorporates both T2I and IT2I generation.

\noindent \textbf{AI-generated Image Detection.} The subtle imperfections in AI-generated images often escape human detection \cite{bray2023testing}. However, they may be identified through sophisticated techniques such as edge detectors, quality metrics, and frequency analysis  \cite{agarwal2017swapped,akhtar2019comparative,mo2018fake,zhang2017automated}. Furthermore, invisible signatures such as camera CFA patterns have been exploited to differentiate camera-generated images from AI-generated images~\cite{reidyCFA2023}. Additionally, anomalies such as improper alignment with the rest of the image \cite{li2018exposing,li2020celeb}, inconsistent lighting \cite{straub2019using}, and differences in image fidelity \cite{korshunov2018deepfakes} also aid in recognition. Systems in \cite{wang2020cnn,reidyCFA2023} finetuned pre-trained models such as ResNet-50 \cite{he2016deep} and ConvNext-S \cite{liu2022convnet}. \cite{ojha2023towards} trains the last linear layer of CLIP-ViT-L \cite{radford2021learning}. \cite{xi2023ai} developed a network consisting of both residual and content features to capture textural differences, particularly in low-frequency areas. In Section \ref{sec:bench}, we show that the state-of-the-art AI image detectors provide unsatisfactory performance on the ARIA dataset. 

The study in \cite{ha2024organic}, independently performed from our work, is the most similar to this project. They collected a dataset for (copyrighted) artworks and tested the identifiability with automatic detectors and human evaluators. Our work is more diverse and comprehensive in the following aspects: (1) their dataset solely covers art style imitation, which is one of our three distinct attack scenarios. (2) The ARIA dataset will be openly shared with the research community for future AIGC research, while the dataset in \cite{ha2024organic} contains proprietary artworks that are unlikely to be shared. (3) Our dataset, ARIA, is significantly larger with 144K images compared to 630 in \cite{ha2024organic}. The size of the ARIA dataset will facilitate the adoption of DNNs in AI-art detection. (4) We have recruited a highly diverse group of participants in the user study, which better represents the range of potential real-world victims of adversarial AI art.

\section{Conclusion}\label{sec:con} 
In this paper, we make a systematic attempt at understanding and detecting adversarial AI art. We first introduce the ARIA dataset, which contains over 120,000 AI images categorized into five categories representing three distinct attack scenarios. We present a user study to evaluate if human users can distinguish between human-generated and AI-generated images. 
We further present a benchmark analysis of open-source and commercial AI detectors, together with a ResNet model trained from scratch using the ARIA dataset. The findings reveal significant challenges for both humans and AI systems in accurately identifying AI-generated content, underscoring the need for advanced strategies to cope with the potential risks introduced by Generative AI.
 
%
%
%
\bibliographystyle{splncs04}
\bibliography{ref}

\begin{thebibliography}{100}
\providecommand{\url}[1]{\texttt{#1}}
\providecommand{\urlprefix}{URL }
\providecommand{\doi}[1]{https://doi.org/#1}

\bibitem{10million}
10 midjourney statistics demonstrating why its better than other ai art generators. Skim AI (2024)

\bibitem{aigovern2}
Risk in focus: Generative a.i. and the 2024 election cycle. CISA (2024)

\bibitem{agarwal2017swapped}
Agarwal, A., Singh, R., Vatsa, M., Noore, A.: Swapped! digital face presentation attack detection via weighted local magnitude pattern. In: IEEE IJCB (2017)

\bibitem{akhtar2019comparative}
Akhtar, Z., Dasgupta, D.: A comparative evaluation of local feature descriptors for deepfakes detection. In: IEEE HST (2019)

\bibitem{n24news}
billywzh717: N24news. \url{https://github.com/billywzh717/N24News} (2022)

\bibitem{bird2024cifake}
Bird, J.J., Lotfi, A.: Cifake: Image classification and explainable identification of ai-generated synthetic images. IEEE Access  (2024)

\bibitem{bray2023testing}
Bray, S.D., Johnson, S.D., Kleinberg, B.: Testing human ability to detect ‘deepfake’images of human faces. Journal of Cybersecurity  \textbf{9}(1),  tyad011 (2023)

\bibitem{brown2020language}
Brown, T., Mann, B., et~al.: Language models are few-shot learners. NeurIPS  (2020)

\bibitem{chan2024one}
Chan, K., Swenson, A.: One tech tip: How to spot ai-generated deepfake images. The Associated Press (2024)

\bibitem{chen2021evaluating}
Chen, M., Tworek, J., et~al.: Evaluating large language models trained on code. arXiv:2107.03374  (2021)

\bibitem{lawsuit}
Cho, W.: Ai companies take hit as judge says artists have ‘public interest’ in pursuing lawsuits. ARTnews (2024)

\bibitem{croitoru2023diffusion}
Croitoru, F.A., Hondru, V., Ionescu, R.T., Shah, M.: Diffusion models in vision: A survey. IEEE TPAMI  (2023)

\bibitem{dintentdata2024ai}
Data, E.D.I.: Case 1739 ai-generated image showing accident between gmc hummer ev and tesla cybertruck has gone viral with false claims. D-Intent Data (2024)

\bibitem{deng2009imagenet}
Deng, J., Dong, W., Socher, R., Li, L.J., Li, K., Fei-Fei, L.: Imagenet: A large-scale hierarchical image database. In: CVPR. Ieee (2009)

\bibitem{dhariwal2021diffusion}
Dhariwal, P., Nichol, A.: Diffusion models beat gans on image synthesis. NeurIPS  (2021)

\bibitem{Duffy_2024}
Duffy, C.: Top ai photo generators produce misleading election-related images, study finds. CNN (2024)

\bibitem{banai}
Edwards, B.: Flooded with ai-generated images, some art communities ban them completely. arstechnica (2022)

\bibitem{epstein2020gets}
Epstein, Z., Levine, S., Rand, D.G., Rahwan, I.: Who gets credit for ai-generated art? Iscience  \textbf{23}(9) (2020)

\bibitem{pixiv2019dataset}
Evan, S.: Pixiv top daily illustration 2018. Kaggle, \url{https://www.kaggle.com/datasets/stevenevan99/pixiv-top-daily-illustration-2018} (2019)

\bibitem{fakeimgdetector}
{Fake Image Detector}: Fake image detector. \url{https://www.fakeimagedetector.com/contact-us/} (2024)

\bibitem{goodfellow2014generative}
Goodfellow, I., Pouget-Abadie, J., Mirza, M., Xu, B., Warde-Farley, D., Ozair, S., Courville, A., Bengio, Y.: Generative adversarial nets. NeurIPS  (2014)

\bibitem{gragnaniello2021gan}
Gragnaniello, D., Cozzolino, D., Marra, F., Poggi, G., Verdoliva, L.: Are gan generated images easy to detect? a critical analysis of the state-of-the-art. In: IEEE ICME (2021)

\bibitem{ha2024organic}
Ha, A.Y.J., Passananti, J., Bhaskar, R., Shan, S., Southen, R., Zheng, H., Zhao, B.Y.: Organic or diffused: Can we distinguish human art from ai-generated images? arXiv:2402.03214  (2024)

\bibitem{Nahrawy2023aiornot}
{Hassan Hicham ElNahrawy}: Ai or not. \url{https://huggingface.co/Nahrawy/AIorNot/} (2023)

\bibitem{he2016deep}
He, K., Zhang, X., Ren, S., Sun, J.: Deep residual learning for image recognition. In: CVPR (2016)

\bibitem{he2021beyond}
He, Y., Yu, N., Keuper, M., Fritz, M.: Beyond the spectrum: Detecting deepfakes via re-synthesis. arXiv:2105.14376  (2021)

\bibitem{ho2020denoising}
Ho, J., Jain, A., Abbeel, P.: Denoising diffusion probabilistic models. NeurIPS  (2020)

\bibitem{hong2024wildfake}
Hong, Y., Zhang, J.: Wildfake: A large-scale challenging dataset for ai-generated images detection. arXiv:2402.11843  (2024)

\bibitem{art2023dataset}
Icaro: Best artworks of all time. Kaggle, \url{https://www.kaggle.com/datasets/ikarus777/best-artworks-of-all-time} (2023)

\bibitem{isitai}
{isitai.com}: Is it ai? \url{https://isitai.com/} (2024)

\bibitem{karras2018progressive}
Karras, T., Aila, T., Laine, S., Lehtinen, J.: Progressive growing of gans for improved quality, stability, and variation. In: ICLR (2018)

\bibitem{survey_ethics}
Katatikarn, J.: Ai art statistics: The ultimate list in 2024. Academy Of Animated Art (2024)

\bibitem{Kavafian_2023}
Kavafian, H.: How to identify ai-generated images (2023)

\bibitem{illuminarty}
{KI-Tech Hertig}: Illuminarty. \url{https://illuminarty.ai/} (2024)

\bibitem{korshunov2018deepfakes}
Korshunov, P., Marcel, S.: Deepfakes: a new threat to face recognition? assessment and detection. arXiv:1812.08685  (2018)

\bibitem{instany100k}
koushikvikram: Multimodal image retrieval. Github, \url{https://github.com/koushikvikram/multimodal-image-retrieval} (2021)

\bibitem{krizhevsky2009learning}
Krizhevsky, A., Hinton, G., et~al.: Learning multiple layers of features from tiny images  (2009)

\bibitem{lei2023rgbd2}
Lei, J., Tang, J., Jia, K.: Rgbd2: Generative scene synthesis via incremental view inpainting using rgbd diffusion models. In: CVPR (2023)

\bibitem{li2018exposing}
Li, Y., Lyu, S.: Exposing deepfake videos by detecting face warping artifacts. arXiv:1811.00656  (2018)

\bibitem{li2020celeb}
Li, Y., Yang, X., Sun, P., Qi, H., Lyu, S.: Celeb-df: A large-scale challenging dataset for deepfake forensics. In: CVPR (2020)

\bibitem{lin2014microsoft}
Lin, T.Y., Maire, M., Belongie, S., Hays, J., Perona, P., Ramanan, D., Doll{\'a}r, P., Zitnick, C.L.: Microsoft coco: Common objects in context. In: ECCV (2014)

\bibitem{liu2023check}
Liu, Z., Yao, Z., Li, F., Luo, B.: On the detectability of chatgpt content: Benchmarking, methodology, and evaluation through the lens of academic writing. arXiv:2306.05524  (2023)

\bibitem{liu2022convnet}
Liu, Z., Mao, H., Wu, C.Y., Feichtenhofer, C., Darrell, T., Xie, S.: A convnet for the 2020s. In: CVPR (2022)

\bibitem{lu2024seeing}
Lu, Z., Huang, D., Bai, L., Qu, J., Wu, C., Liu, X., Ouyang, W.: Seeing is not always believing: Benchmarking human and model perception of ai-generated images. NeurIPS  (2024)

\bibitem{maybe2024detector}
{Matthew Maybe}: Ai image detector. \url{https://huggingface.co/umm-maybe/AI-image-detector/} (2022)

\bibitem{insta2022dataset}
Matusevski, A.: Instagram images - 1,211,625 posts. Kaggle, \url{https://www.kaggle.com/datasets/shmalex/instagram-images} (2022)

\bibitem{midj2024copyright}
Midjourney: Can i use my images commercially? MidJourney, \url{https://help.midjourney.com/en/articles/8150363-can-i-use-my-images-commercially} (2024)

\bibitem{midjourney2024home}
{Midjourney}: Midjourney home. \url{https://www.midjourney.com/home} (2024)

\bibitem{aielection1}
Mirza, R.: How ai deepfakes threaten the 2024 elections. The Journalist’s Resource (2023)

\bibitem{mo2018fake}
Mo, H., Chen, B., Luo, W.: Fake faces identification via convolutional neural network. In: ACM workshop on information hiding and multimedia security (2018)

\bibitem{nichol2021improved}
Nichol, A.Q., Dhariwal, P.: Improved denoising diffusion probabilistic models. In: ICML (2021)

\bibitem{usco}
Office, U.S.C.: Usco letter on ai and copyright initiative update  (2024)

\bibitem{ojha2023towards}
Ojha, U., Li, Y., Lee, Y.J.: Towards universal fake image detectors that generalize across generative models. In: CVPR (2023)

\bibitem{dalle2024copyright}
OpenAI: Can i sell images i create with dall·e? OpenAI Documentation (2024)

\bibitem{openai2024dalle}
{OpenAI}: Dall·e: Creating images from text. \url{https://openai.com/research/dall-e} (2024)

\bibitem{organika2024sdxl}
{Organika.ai}: Sdxl detector. \url{https://huggingface.co/Organika/sdxl-detector/} (2024)

\bibitem{deepfakedating}
Pashankar, S.: Scammers litter dating apps with ai-generated profile pics. Bloomberg (2024)

\bibitem{podell2023sdxl}
Podell, D., English, Z., Lacey, K., Blattmann, A., Dockhorn, T., M{\"u}ller, J., Penna, J., Rombach, R.: Sdxl: Improving latent diffusion models for high-resolution image synthesis. arXiv:2307.01952  (2023)

\bibitem{rabiner1986introduction}
Rabiner, L., Juang, B.: An introduction to hidden markov models. IEEE ASSP Magazine  \textbf{3}(1),  4--16 (1986)

\bibitem{radford2021learning}
Radford, A., Kim, J.W., et~al.: Learning transferable visual models from natural language supervision. In: ICML (2021)

\bibitem{rae2021scaling}
Rae, J.W., Borgeaud, S., et~al.: Scaling language models: Methods, analysis \& insights from training gopher. arXiv:2112.11446  (2021)

\bibitem{rahman2023artifact}
Rahman, M.A., Paul, B., Sarker, N.H., Hakim, Z.I.A., Fattah, S.A.: Artifact: A large-scale dataset with artificial and factual images for generalizable and robust synthetic image detection. In: IEEE ICIP (2023)

\bibitem{ramesh2022hierarchical}
Ramesh, A., Dhariwal, P., Nichol, A., Chu, C., Chen, M.: Hierarchical text-conditional image generation with clip latents. arXiv:2204.06125  (2022)

\bibitem{reidyCFA2023}
Reidy, M., Mallon, H., Luo, J.: Investigating the effectiveness of deep learning and {CFA} interpolation based classifiers on identifying {AIGC}. In: IEEE BigData (2023)

\bibitem{reynolds2009gaussian}
Reynolds, D.A., et~al.: Gaussian mixture models. Encyclopedia of biometrics  \textbf{741}(659-663) (2009)

\bibitem{rombach2022high}
Rombach, R., Blattmann, A., Lorenz, D., Esser, P., Ommer, B.: High-resolution image synthesis with latent diffusion models. In: CVPR (2022)

\bibitem{roose2022ai}
Roose, K.: An ai-generated picture won an art prize. artists aren't happy.  (2022)

\bibitem{rossler2019faceforensics++}
Rossler, A., Cozzolino, D., Verdoliva, L., Riess, C., Thies, J., Nie{\ss}ner, M.: Faceforensics++: Learning to detect manipulated facial images. In: ICCV (2019)

\bibitem{wvolf2024cnn}
{Rudolf Kenechukwu Enyimba}: Deepfake image detection(cnn). \url{https://huggingface.co/spaces/Wvolf/CNN_Deepfake_Image_Detection/} (2024)

\bibitem{deepfakesocial}
Sganga, N.: Is that facebook account real? meta reports 'rapid rise' in ai-generated profile pictures. CBS News (2022)

\bibitem{sha2023fake}
Sha, Z., Li, Z., Yu, N., Zhang, Y.: De-fake: Detection and attribution of fake images generated by text-to-image generation models. In: ACM CCS (2023)

\bibitem{shang2024resdiff}
Shang, S., Shan, Z., Liu, G., Wang, L., Wang, X., Zhang, Z., Zhang, J.: Resdiff: Combining cnn and diffusion model for image super-resolution. In: AAAI (2024)

\bibitem{sightengine}
{Sightengine}: sightengine. \url{https://sightengine.com/} (2024)

\bibitem{sohl2015deep}
Sohl-Dickstein, J., Weiss, E., Maheswaranathan, N., Ganguli, S.: Deep unsupervised learning using nonequilibrium thermodynamics. In: ICML (2015)

\bibitem{song2021maximum}
Song, Y., Durkan, C., Murray, I., Ermon, S.: Maximum likelihood training of score-based diffusion models. NeurIPS  (2021)

\bibitem{song2020improved}
Song, Y., Ermon, S.: Improved techniques for training score-based generative models. NeurIPS  (2020)

\bibitem{dreamstudio2024}
{Stability AI Ltd}: Dreamstudio. \url{https://dreamstudio.com/about/} (2024)

\bibitem{starryai2024home}
{StarryAI}: Starryai home. \url{https://starryai.com/} (2024)

\bibitem{Steele_2024}
Steele, C.: How to detect ai-generated images (2024)

\bibitem{straub2019using}
Straub, J.: Using subject face brightness assessment to detect ‘deep fakes’(conference presentation). In: Real-Time Image Processing and Deep Learning. vol. 10996, p. 109960H. SPIE (2019)

\bibitem{takagi2023high}
Takagi, Y., Nishimoto, S.: High-resolution image reconstruction with latent diffusion models from human brain activity. In: CVPR (2023)

\bibitem{disaster2012dataset}
Telperion: Diasterdatasetraw. Kaggle, \url{https://www.kaggle.com/datasets/telperion/diasterdatasetraw} (2022)

\bibitem{Thompson_2024}
Thompson, S.A.: We asked a.i. to create the joker. it generated a copyrighted image. (2024)

\bibitem{vaswani2017attention}
Vaswani, A., Shazeer, N., Parmar, N., Uszkoreit, J., Jones, L., Gomez, A.N., Kaiser, {\L}., Polosukhin, I.: Attention is all you need. NeurIPS  (2017)

\bibitem{aifakenews1000}
Verma, P.: The rise of ai fake news is creating a ‘misinformation superspreader’. The Washington Post (2023)

\bibitem{wang2020cnn}
Wang, S.Y., Wang, O., Zhang, R., Owens, A., Efros, A.A.: Cnn-generated images are surprisingly easy to spot... for now. In: CVPR (2020)

\bibitem{wang2023benchmarking}
Wang, Y., Huang, Z., Hong, X.: Benchmarking deepart detection. arXiv:2302.14475  (2023)

\bibitem{wang2022n24news}
Wang, Z., Shan, X., Zhang, X., Yang, J.: N24news: A new dataset for multimodal news classification. In: LREC (2022)

\bibitem{wang2023dire}
Wang, Z., Bao, J., Zhou, W., Wang, W., Hu, H., Chen, H., Li, H.: Dire for diffusion-generated image detection. In: ICCV (2023)

\bibitem{wang2023diffusiondb}
Wang, Z., Montoya, E., Munechka, D., Yang, H., Hoover, B., Chau, P.: Diffusiondb: A large-scale prompt gallery dataset for text-to-image generative models. In: ACL (2023)

\bibitem{wei2024understanding}
Wei, Y., Tyson, G.: Understanding the impact of ai generated content on social media: The pixiv case (2024)

\bibitem{wendling2024ai}
Wendling, M.: Ai can be easily used to make fake election photos. BBC (2024)

\bibitem{wong2024ai}
Wong, C.: Ai-generated images and video are here: how could they shape research? Nature  (2024)

\bibitem{contentatscale}
{Workado LLC}: Content at scale. \url{https://contentatscale.ai/} (2024)

\bibitem{aielection2}
{World Economic Forum}: Global risks report 2024  (2024)

\bibitem{Writer_2024}
Writer, A.R.: 9 simple ways to detect ai images (with examples) in 2024 (2024)

\bibitem{wu2023ai}
Wu, J., Gan, W., Chen, Z., Wan, S., Lin, H.: Ai-generated content (aigc): A survey. arXiv:2304.06632  (2023)

\bibitem{xi2023ai}
Xi, Z., Huang, W., Wei, K., Luo, W., Zheng, P.: Ai-generated image detection using a cross-attention enhanced dual-stream network. In: APSIPA ASC (2023)

\bibitem{yang2023diffusion}
Yang, L., Zhang, Z., Song, Y., Hong, S., Xu, R., Zhao, Y., Zhang, W., Cui, B., Yang, M.H.: Diffusion models: A comprehensive survey of methods and applications. ACM Computing Surveys  \textbf{56}(4),  1--39 (2023)

\bibitem{yu2015lsun}
Yu, F., Seff, A., Zhang, Y., Song, S., Funkhouser, T., Xiao, J.: Lsun: Construction of a large-scale image dataset using deep learning with humans in the loop. arXiv:1506.03365  (2015)

\bibitem{zhang2017automated}
Zhang, Y., Zheng, L., Thing, V.L.: Automated face swapping and its detection. In: IEEE ICSIP (2017)

\bibitem{zhu2024genimage}
Zhu, M., Chen, H., Yan, Q., Huang, X., Lin, G., Li, W., Tu, Z., Hu, H., Hu, J., Wang, Y.: Genimage: A million-scale benchmark for detecting ai-generated image. NeurIPS  (2024)

\end{thebibliography}

\end{document}